\documentclass[12pt,final,a4paper,oneside]{report}

\usepackage[utf8]{inputenc}
\usepackage{graphicx}
\usepackage{geometry}
\usepackage{pstricks}
\usepackage{setspace}
\usepackage[final]{pdfpages}
\usepackage{subcaption}

\usepackage[hidelinks]{hyperref}
\hypersetup{linktoc=all}

\usepackage{xspace}

\newcommand{\at}{$a_{t}$\xspace}
\newcommand{\st}{$s_{t}$\xspace}
\newcommand{\rt}{$r_{t}$\xspace}
\newcommand{\ot}{$o_{t}$\xspace}
\newcommand{\bigO}{$O$\xspace}
\newcommand{\invbigO}{$O^{-1}$\xspace}

\begin{document}

\begin{titlepage}
    \begin{center}
    \vspace*{1.0cm}{
        \Huge{\textbf{``What’s my model inside of?'': Exploring the role of environments for grounded natural language understanding}\par}
    }
    
    \begin{doublespacing}
    
    \vspace*{1.5cm} {\Large{Thesis for the degree of \\ ``Doctor of Philosophy''}}
    
    \vspace*{1.5cm} {\Large{by \\ Ronen Tamari}}
    
    \vspace*{5.5cm} {\large{Submitted to the Senate of the Hebrew University of Jerusalem\\ October 2023}}
    
    
    \end{doublespacing}
    \end{center}

\end{titlepage}

\pagestyle{empty} 

\newpage
\mbox{}
\newpage

\begin{doublespacing}
\begin{center}
\vspace*{3.0cm}  {\Large {This work was carried out under the supervision of \\}}
\vspace{0.3cm}
{\Large {Prof. Dafna Shahaf and Prof. Reut Tsarfaty\\}}
\end{center}
\end{doublespacing}

\newpage
\mbox{}
\newpage

\section*{Acknowledgments}

First, I would like to thank my advisors, Prof. Dafna Shahaf and Prof. Reut Tsarfaty. Dafna provided sharp and constructive criticism as well as patience and support. Both aspects have been instrumental to my academic growth. I am especially appreciative of her willingness to let me follow my heart on far-ranging explorations, and the trust she placed in me. She also inspired me with her super-human clarity during 3am paper writing sessions before submission deadlines. Reut opened my mind to the intriguing concept of semantics, and relations between natural and formal languages. She connected me to mainstream NLP and encouraged me to pay attention to up-and-coming pre-trained models while I was pre-occupied with other topics, and this turned out to be fateful advice. She quite literally helped me ground my thinking around grounding in natural language, and I am grateful to her for her support and ``yes, and'' approach to research. I am very fortunate to have had the guidance of both Dafna and Reut throughout my PhD. 

I am grateful for the support of my committee, Dr. Gabriel (Gabi) Stanovsky and Jason Baldridge. Beyond a committee member, Gabi mentored me during my first internship at the Allen Institute for Artificial Intelligence in Seattle. Gabi taught me to appreciate clarity and simplicity and showed me how to apply the KISS principle in my research. He helped me advance my research far beyond where I could take it alone, and connected me with Prof. Alan Ritter and his PhD student Fan Bai at Georgia Tech, who I am also grateful to have collaborated with. Gabi was very generous to me in and out of the lab, and inspired me to try and pay it forward. Jason was a great inspiration to me as I started my PhD journey. Watching his fantastic keynote at the 2019 NeurIPS conference was a formative experience that deeply shaped my research interests. I especially appreciated his inter-disciplinary approach and ability to skillfully integrate linguistics and cognitive science research with complex computational modelling and engineering challenges.

I also want to extend a special thanks to Dr. Kyle Richardson, who was not officially on my PhD committee but very well could have been - he was also a friend and mentor who I met in Seattle and later collaborated with. From Kyle I learned what it takes to build NLP pipelines at industrial scale and precision, and I would not have been able to develop and run experiments on large pre-trained models without his significant help.  Kyle invested significant time in helping me develop my ideas at both the conceptual and empirical level and I benefited greatly from being able to work closely with him. 

I feel very fortunate to have worked with a wide group of collaborators at Hebrew University, especially Dr. Tom Hope, Chen Shani and Oren Sultan. Thanks to Prof. Omri Abend for helping me in my first orientation steps in cognitive science and linguistics. Thanks to all of Dafna Shahaf's Hyadata Lab, which was my home away from home. Indeed, during some stretches of my PhD, I spent more time there than at home. Thanks especially to lab-mates Tom, Chen and Moran Mizrahi for their friendship and inspiration.

I was also very fortunate to begin my PhD journey in a summer internship at the RIKEN Institute in Japan, working with Prof. Yuji Matsumoto at the Nara Institute of Technology (NAIST). In Japan I learnt the haiku ``distant minds meet, cherries blossom,'' which describes well my time there and since. Thanks to Prof. Matsumoto for providing me the generous support, peaceful setting and challenging problem which sowed the seeds of inspiration for the rest of my PhD.

My experiments around the Dyna-bAbI benchmark required large-scale engineering and experimentation efforts I would not have been able to undertake alone. Thanks to Noam Kahlon and Aviad Sar Shalom who volunteered time and effort to help me build the Dyna-bAbI data generator. Thanks also to Nelson Liu at Stanford who pitched in to help with running experiments on the new dataset we created.

Thanks to Prof. Judy Fan and her wonderful cogtools lab at the University of California in San Diego (now Stanford). Judy hosted me generously for an exchange program, and I deeply appreciated the opportunity to gain a cognitive science perspective on NLP. 

For the last phase of my thesis, I am very grateful to have partnered with William Fischer and Lauren Hebert from Veeo who inspired me to go all in on collective sensemaking. I am also grateful to Matan Field from DAOstack who supported me and connected me with the wonderful, wild world of decentralized collaboration. Special thanks to Daniel Friedman at UC Davis who introduced me to the mind-expanding concept of stigmergy and who has continued to expand my mind at a regular basis since then. With the closing of my thesis I have gained new appreciation and wonder at the truly collective nature of intelligence, but a side effect is a feeling of frustration with the incompleteness of acknowledgements. We are but individual ``neurons'' often pitifully unaware of our role in the larger collective network. Thanks to all those many friends, colleagues and collaborators who have been a part of this network and supported me throughout.

A special thanks to my close and extended family who have supported me faithfully through what at times seemed a never-ending journey. To my parents Cathy and Yoav, who reminded me to get out and breathe fresh air once in a while. To my brother Natan, who got me hooked on simulations and multi-agent systems (SimCity, Civilization, etc) somewhere around preschool. Finally, this thesis owes its existence due to my wife and love of my life, Shiran. We met during an internship in Seattle, and my life has changed forever since, and with it my thesis too. I can't imagine either without you.



\newpage
\mbox{}
\newpage

\pagestyle{plain} 
\pagenumbering{roman}

\section*{Abstract}
In recent years, deep learning based approaches to natural language processing (NLP) have made impressive progress. A particularly important achievement has been the development of Large Language Models (LLMs), massive artificial neural networks trained on internet scale linguistic data. LLMs have demonstrated remarkable performance across a wide variety of tasks, including long standing challenges like few-shot learning and coherent long-form text generation.

LLMs seem to be doing more than just ``processing'' natural language (NLP), perhaps they are also \emph{understanding} natural language (NLU)? Indeed, a core ongoing debate being fueled by these advances is the Symbol Grounding Problem; computational models of language process only linguistic input (symbols), so how can their outputs be grounded to the external world to which the language refers to? Or in other words, since meaning also includes extra-linguistic referents (actions, percepts, semantic knowledge), can LLMs reliably understand language, i.e., extract the meaning conveyed by linguistic symbols? The debate is far from settled, and has significant implications for guiding the future of NLP research as well as real-world applications; some are claiming LLM research is climbing the wrong hill altogether, some advocate for more cognitively inspired architectures, while others believe that LLMs are early demonstrations of so-called ``Artificial General Intelligence''.

This PhD thesis proposes a novel environment-oriented perspective on the language grounding debate and towards NLU research more broadly. Our approach is inspired by ecological accounts of cognition, which provide a more holistic account of the role of the body and environment in shaping cognition, in contrast to classical cognitive science which focused on studying brains in isolation. Similarly, we adopt an ecological approach to NLU research. Where current NLU research tends to focus on models in isolation, we developed a more holistic perspective accounting for the deep yet under-explored coupling between models and the environments with which they interact. Wittgenstein famously wrote that “The limits of my language mean the limits of my world”. The thesis can be stated simply, in an inversion of that dictum, that “The limits of my environments mean the limits of my language (models)”. Environments are the computational spaces in which models are embedded, which support data annotation as well as model training, development and evaluation. The ecological perspective provides novel insights on each of those critical components of the NLU research pipeline, and also contributes to the broader debate around how to pursue more reliable and grounded natural language understanding systems. 

The thesis is divided into three parts. The first part synthesizes research from cognitive science, linguistics, reinforcement learning and NLU to propose a conceptual framework for the idea of ecological natural language understanding. The framework highlights metaphor comprehension, world model learning and mental simulation as core capacities of importance for achieving human-like natural language understanding. The framework makes predictions about the limitations of statistical language models (such as LLMs) with respect to these abilities, and also about the limitations of benchmarks used to evaluate them. Importantly, the ecological NLU approach suggests that many debates about language grounding are hamstrung by the lack of appropriate benchmarks for evaluation of language understanding, both in terms of complexity and rigor. The framework is used to outline an NLU research roadmap for addressing the described limitations.

The second part of the thesis begins advancing on the roadmap described in the first section, applying the theoretical framework towards more practical applications. We developed text-based game environments supporting novel training and annotation methods for procedural text understanding. We found that the more detailed annotations made possible using text based games enabled more faithful modelling of process-level comprehension tasks, compared with existing approaches that addressed mainly sentence-level comprehension. We also used text-based games to construct a new benchmark for measuring the progress of language models on challenging commonsense reasoning tasks. We used the benchmark to identify limitations of state-of-the-art models in tasks requiring compositional generalization, and explored data augmentation strategies to improve models' generalization. Finally, we leveraged the richer supervision provided by text-based game environments to develop Breakpoint Transformers, an extension of the Transformer architecture designed to model intermediate semantic information in long narrative or procedural texts. We applied Breakpoint Transformers on a challenging common-sense reasoning task and achieved significant improvements over existing approaches, both in terms of accuracy (close to 300\% performance increase in the primary sub-task) as well as architecture generality.

In the third and final part of the thesis, we explored the implications of ecological cognition for the design of online epistemic environments for humans. If indeed environments play such an integral role in human (and machine) cognition, perhaps many of the “epistemic ills” plaguing societies (e.g., polarization, misinformation) today can be traced back to the online social media environments from which humans increasingly acquire their information about the world? What would healthier epistemic environments look like? We integrated Semantic Web research with theories from epistemology and distributed cognition to provide a novel, ecological perspective, on collective intelligence (or lack thereof). We highlighted risks inherent to current centralized social media platforms and proposed a new kind of epistemic environment addressing their limitations, in the form of decentralized, AI-augmented collective intelligence networks.
\newpage

\newpage

\tableofcontents

\newpage
\mbox{}
\newpage

\pagenumbering{arabic}
\chapter{Introduction}
\label{chap:Introduction}

 \begin{quotation}
\noindent ``Ask not what's inside your head, but what your head's inside of.'' -- W.M. Mace
 \end{quotation}

In recent years, deep learning based approaches to natural language processing (NLP) have made impressive progress. A particularly important achievement has been the development of Large Language Models (LLMs), massive artificial neural networks trained on internet scale linguistic data. LLMs have demonstrated remarkable performance across a wide variety of tasks, including long standing challenges like few-shot learning and coherent dialog and long-form text generation~\cite{brown2020language,glaese2022improving,openai2023gpt4}.

The success of LLMs has significant practical and theoretical implications. On the practical side, LLMs are increasingly capable of performing natural language “understanding” (NLU) tasks, beyond merely natural language processing (NLP). We use “understanding” to distinguish between human-like understanding and the more technical usage of the term in computational linguistics\footnote{See also Melanie Mitchell on the use of ``wishful mnemonics'' by AI researchers~\cite{mitchell2021ai}.}; NLP tasks traditionally focus on simpler supporting operations like part-of-speech tagging or entity recognition, whereas NLU tasks involve extracting some form of meaning from an utterance (e.g., mapping a natural language question to a formal database query)~\cite{bender2022linguistic}. On the theoretical side, LLMs' success has fueled long-standing debates about the nature of linguistic meaning~\cite{mollo2023vector}, leading many cognitive scientists and linguists to update their theories of language production and understanding.

These advances also raise pressing questions. For practical NLU applications, despite their impressive performance, LLMs still display a stubborn tendency for commonsense reasoning failures and hallucination of fake facts~\cite{zhang2023sirens,gendron2023large,substackLargeLanguage}. As observed by John Oliver, ``The problem with AI isn't that it's smart, but that it is dumb in ways we can't predict.''\footnote{\url{https://www.youtube.com/watch?v=Sqa8Zo2XWc4}} Can we get better at predicting the limitations of LLMs? In what settings can they be safely deployed, and how can their reliability be improved? These more practical questions are in turn informed by more philosophical and theoretical questions: what is linguistic meaning, and how is meaning conveyed by linguistic utterances? To what degree is meaning reducible to statistical patterns in language? What does it mean to understand language, and are machines capable of human-like language understanding? What model architectures might achieve more human-like NLU capabilities?

Our thesis contributes to both the practical and theoretical questions, by introducing a novel, \emph{ecological} perspective on natural language understanding. Our approach is inspired by ecological accounts of cognition and linguistics. Common to those theories is a more holistic account of the role of the environment in shaping cognition and language, in contrast to classical cognitive science which focused on studying brains in isolation. We observe a similar parallel in NLU research, where most of the focus in mainstream research is on modelling artifical neural networks, and environments fade unnoticed into the background. In our thesis, we ask what happens when we foreground environments, and treat them as ``first-class research citizens'' alongside models. In particular, the thesis addresses the following research questions:

\begin{itemize}
    \item Chapter \ref{chap:lang_remodel}: how can we account for the role of the environment in a computational framework of natural language understanding?
    \item Chapters \ref{chap:ptb},\ref{chap:scientific_protocols}: how can environments inform more realistic training, evaluation and annotation methods in procedural text understanding tasks? How can such environments be created efficiently?
    \item Chapter \ref{chap:dynababi}: what are the world modelling capabilities of neural language models, and how can we judge if a world modelling benchmark is effective?
    \item Chapter \ref{chap:breakpoint_transformers}: how can existing LLM architectures be extended to better incorporate world-modelling capabilities without harming existing performance or compromising on computational efficiency?
    \item Chapter \ref{chap:stigmergic_annotation}: how can networked annotation tools support AI-augmented collective (human) sensemaking environments, as a healthier and more open alternative to centralized social media platforms with opaque data and algorithms?

    
\end{itemize}

The following sections provide further background on environments to motivate and contextualize these questions: \S\ref{sec:envs} discusses environments in cognitive science, linguistics and reinforcement learning, as well as from a meta-science perspective asking how environments are seen by AI researchers. \S\ref{sec:eco_nlu} discusses how the thesis translates those accounts translate into current NLU research, and \S\ref{sec:eco_sense} discusses environments for social learning and sensemaking in humans.

\section{Environments: the cognitive, the computional and the meta}
\label{sec:envs}

\subsection{Environments in ecological cognitive science}
\label{ssec:eco_cog_envs}

The human brain is perhaps the most complex object in the known universe, and as a result, has been the focus of intense research spanning many disciplines, from philosophy through cognitive science, neuroscience, and increasingly also artificial intelligence. A realization emerging across these fields is that brains and cognition cannot be understood in isolation; ``cognition does not occur exclusively inside the head''~\cite{carney2020thinking}. Rather, the explanation of many cognitive phenomena (including language) necessitates a more ecological approach accounting for the role of the body and wider environment.\footnote{We use ecological cognitive science for brevity to refer to a range of approaches sharing the ecological focus, including ecological psychology, 4E-cognition (embodied, embedded, extended and enactive), grounded cognition and embodied cognitive linguistics. See Chapter \ref{chap:lang_remodel} and Appendix \ref{chap:eco_sem} for more background.} While the simple observation that bodies and environments matter may seem trivial at first glance (brain need bodies as an energy supply), the \emph{ways} and \emph{degree} to which they matter often turn out to be surprising, and with radical implications for both theory and methodology. 

\subsubsection{Outfielder Problem: an example of ecological cognition in action}
An illustrative example is the ``outfielder problem:''~\cite{Fink2009,Kelty_2022}  how does a baseball outfielder catch a fly ball? A traditional ``in-the-head cognition'' approach would attempt to solve the problem assuming that all that was available to solve the problem was the brain. It would break the problem into subproblems corresponding to cognitive modules implemented by the brain: a sensor module to detect the ball, an inference module to calculate its trajectory given a physics world model, and an action planner to produce and execute a motor plan to bring the player to the predicted landing site. Such an approach is plausible on the surface, but requires highly accurate measurements that are difficult to estimate in practice. Ecological cognition takes a radically different perspective (which turns out to be an accurate account for how real baseball players solve the problem), and asks - what if the body and the environment were part of the solution? In baseball terms: as long as you keep your eye on the ball and move to keep your gaze steady, then the ball will fall into a glove brought in front of your face. The outfielder is leveraging rich and constant feedback from the environment, as well as the ability to act all the while: simply moving so as to make the world appear a certain way now, which leads them to be in the right place later. This example demonstrates how ecological cognition embodies (literally) a very different approach both in terms of theory and methodology; imagine how different a robot outfielder design would look whether informed by ecological or traditional cognitive science.

\subsection{Environments in cognitive linguistics}

The field of linguistics has also been deeply impacted by ecological cognitive theories. George Lakoff's work~\cite{lakoff1980metaphorical,lakoff2008metaphors,dodge2005image}, particularly in cognitive linguistics and conceptual metaphor theory, has been instrumental in this transformation. Lakoff argued that language is deeply rooted in our sensory experiences and bodily interactions with the world. His exploration of metaphors, such as "time is money" or "love is a journey," illustrated how abstract concepts are pervasively structured by our embodied experiences, highlighting the fundamental role of the body and environment in shaping linguistic expressions. These theories ushered in a paradigm shift that has challenged the traditional views advocated by cognitive scientists like Noam Chomsky and Jerry Fodor, who championed more innate and abstract approaches to language that did not account for the role of the environment~\cite{harris_linguistics_2021}.

Of particular interest in the context of NLU research, environments feature centrally in the symbol grounding problem, a central debate in AI and cognitive science~\cite{HARNAD1990335,mollo2023vector}. The debate seeks to understand how systems processing only symbols, such as words, come to acquire the meaning, or semantics, of those symbols. Ecological theories of cognition contend that language is grounded through its referring to an extra-linguistic world of objects, actions, events and semantic knowledge. The environment thus provides the context and sensory input that allows symbols to become grounded in real-world experiences. These theories raise significant questions for NLU research: computational models of language exposed only to linguistic data have no direct experience of the world, thus it is unclear whether they can reliably connect between linguistic inputs to the world the language refers to. The degree to which environments are necessary is still an open question: theories of indirect grounding~\cite{harnad_2016} suggest that perhaps only some part of language must be grounded directly to sensorimotor experience, while other language can be grounded indirectly through this base.

\subsection{Environments in reinforcement learning}

Reinforcement Learning (RL) is one of the branches of AI in which environments play the most prominent role. Indeed, RL is foundational to the sub-field of embodied AI~\cite{duan2022survey}, which focuses on developing agents which can solve tasks involving interaction in real (e.g., production line robots) or virtual environments (e.g., a web browsing agent). As such, RL is a natural starting point for a formal computational account of environments. The standard formulation of RL is given by the Partially Observable Markov Decision Process~\cite{kaelbling1998planning}, which describes the agent-environment interaction.

\begin{figure}
    \centering
    \includegraphics[width=1\textwidth]{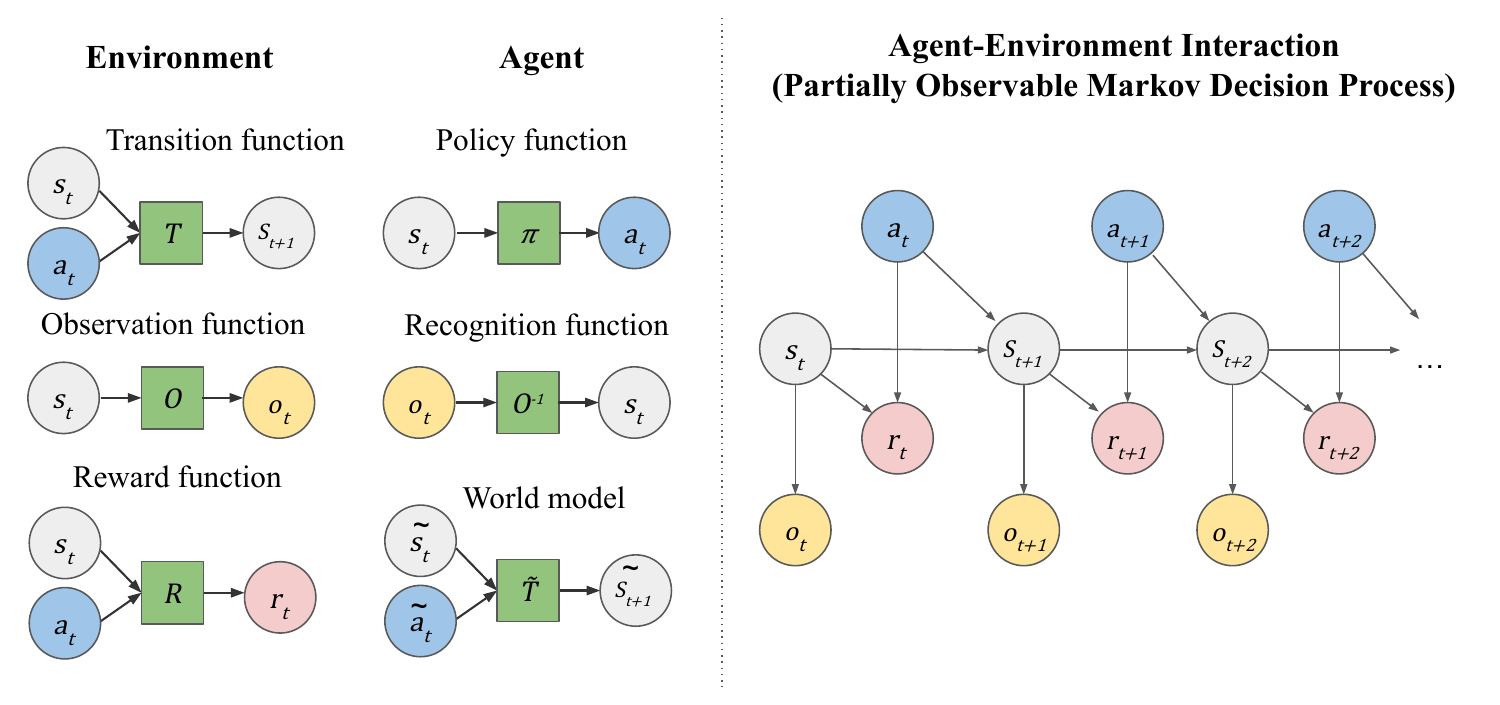}
    \caption{Agent-environment interaction dynamics, modelled by a Partially Observable Markov Decision Process (POMDP). Figure based on~\cite{Hamrick_2019}.}
    \label{fig:agent_env}
\end{figure}

As can be seen from Figure \ref{fig:agent_env}, the computational core of the environment is the transition function, $T$. The transition function represents the world dynamics: how will the world change given a particular action \at in state \st. In partially observable environments, agents may not have access to the full world state; the observation function \bigO is the process by which sensory data (the observation the agent has access to) are generated from the world state \st. For example, in a 3D computer game, \st will contain the full scene specification (objects, locations, physical forces, etc) and \ot will contain a rendering of those data as a frame of pixels representing the visual field of the agent. Finally, environments may also include a reward function $R$, providing agents with a reward signal \rt based on their current action \at and world state \st. Agents interact with the environment by means of a \emph{policy} function, $\pi$, which generates an action \at given the current world state \st. In the case of a partially observable environment, the agent must first employ a \emph{recognition} function \invbigO to parse the observation \ot into recognizable state \st. Model-based reinforcement learning approaches employ a world model $\tilde{T}$ that is learned through interaction with $T$ and can be used for mental simulation and planning with mental states $\tilde{s}$ and actions $\tilde{a}$~\cite{Hamrick_2019}.

Simulators are a classic kind of environment, used extensively in RL research. Simulators include graphical interfaces (e.g., a 3D game engine) or can be purely text-based (discussed at length in Chapters \ref{chap:lang_remodel},\ref{chap:ptb},\ref{chap:scientific_protocols}). Beyond simulators, environments include any interface that exposes an API (Application Programming Interface) that can be used by an agent.

Note that not all environment functions even need to be implemented to support learning algorithms. The next-word prediction training framework of LLMs can also be framed as an environment, where actions and observations correspond to words, and the learner is exposed to a stream of such observations. There is no meaningful interaction in the sense that the next observation is independent of the learner's action (word prediction). Yet, as discussed in Chapter \ref{chap:dynababi}, even this impoverished environment can drive powerful learning algorithms given enough parameters and data.


\subsection{How are environments seen (or not) in AI research?}

Similar to traditional ``in the head'' cognitive science discussed in \S\ref{ssec:eco_cog_envs}, mainstream AI and NLP research are markedly model-centric ~\cite{Sambasivan2021,rogers-2021-changing}, focusing on topics such as neural architectures, optimization methods and learning algorithms. Interestingly, this focus is not just due to the perceived relative importance of models, but rather has sociological factors as well. Data work, a close relative of environment work, is ``de-glamorised and undervalued''~\cite{Sambasivan2021}; ``everyone wants to do the model work, not the data work,'' despite wide-spread acknowledgement of the critical importance of data work.

Work on environments by AI researchers is additionally complicated by a high engineering overhead, limiting environment development mainly to industrial labs or large-scale academic efforts. Furthermore, from a narrow academic disciplinary lens, environment design falls outside of AI, and is closer to fields such as game design or human-computer interaction (HCI).

Finally, recent years have seen mainstream AI and NLU research drifting apart from other related fields such as cognitive science and neuroscience~\cite{Rooij_Guest_Adolfi_Haan_Kolokolova_Rich_2023,zador2023catalyzing}, possibly leading to under-appreciation of certain insights from those fields. For example, in~\cite{zador2023catalyzing}, an inter-disciplinary group of AI and neuroscience researchers recently proposed updating the Turing Test (seen in AI as a canonical test of NLU) to an ``Embodied Turing Test,'' to better reflect the importance of embodied interaction within an environment as a measure of intelligence.

To summarize, for a variety of reasons, of which many are not scientific (but rather sociological, meta-scientific, etc), environments are not yet seen as ``first class citizens'' in AI or NLU research.

\section{Ecological Natural Language Understanding}
\label{sec:eco_nlu}

 \begin{quotation}
\noindent ``Once you see the boundaries of your environment, they are no longer the boundaries of your environment.''  -- Marshall McLuhan
 \end{quotation}

Primed with the cognitive science background, a computational formulation, and McLuhan, we can begin to see environments everywhere in NLU, and where inattention to them is limiting research progress. The unifying thread of the thesis is the identification of ``environment bottlenecks'' and experimentation with environments to contribute to theoretical and empirical research questions in NLU and AI. 

\subsection{Incorporating environments into NLU theory}

In Chapter \ref{chap:lang_remodel} (with extensions in Appendix \ref{chap:eco_sem}), we revisit the language grounding debate with the ecological lens and propose a new conceptual framework for grounded NLU systems, i.e., systems in which language acquires meaning in the context of an external environment including events, actions, perceptions and the mental models of interlocutors. The standard language modelling formulation used in NLU considers language in isolation: a language model is, at its core, a probabilistic model estimating the likelihood of a future utterance given a history of past utterances. We expanded that formulation to an \emph{embodied} language model, which incorporates the mental models of communicators beyond just the linguistic signals of the standard model. Incorporating mental models thus provides a more realistic computational formulation, and also enables leveraging work on world models from embodied AI (specifically, model-based RL) research, which account for the role of the environment: world models are learned through embodied interaction with the environment ($T$). Language thus acquires meaning by its effect on the (mental) world models of listeners situated in real or virtual environments. Understanding can then be construed as a ``meeting of the minds;'' the speaker and listener come to share a similar mental state, and the speaker is using words (and their knowledge of the listener's mental model) to ``program the mind'' or elicit a desired mental state in the listener.

The theoretical framework above also suggests a roadmap for empirical research, namely, creation of richer and more interactive training environments for grounded NLU models, and using those environments for developing and evaluating world-modelling capabilities of models.


\subsection{Empirical applications: putting environments into practice}
Procedural text understanding constitutes a natural starting point for empirical research; procedures are typically situated in some environment in which the procedure is meant to be executed, for example instructions for furniture assembly or a scientific experiment. Chapters \ref{chap:ptb} and \ref{chap:scientific_protocols} focus on the setting of procedural text understanding for laboratory experimental protocols. Prior work focused mainly on extracting structured graph representations, called action graphs, from unstructured procedural texts~\cite{kulkarni-etal-2018-annotated,mysore2019materials}. In this setting, models are trained to label text spans and relations between them as the nodes and edges of the graph. In Chapter \ref{chap:ptb} we present TextLabs, a novel, interactive approach to action graph extraction, where procedural texts are interpreted as instructions in a text-based lab simulator. Our approach provides a number of advantages compared to the standard action-graph setting: support for interactive training methods like RL, a causal, learnable world-model of lab operations, and the ability to generate synthetic data of controllable complexity. We also implemented a simple RL agent as a sanity check for our environment, and found it could successfully execute simple procedures, while failing to complete longer and more complicated tasks.

In Chapter \ref{chap:scientific_protocols}, we demonstrated a novel use of text-based games as annotation environments. We found that action graphs, while useful for tasks like semantic search, were not sufficient for other important downstream applications, such as parsing of protocols to code for execution in robotic ``cloud labs''~\cite{lee2018autoprotocol,Miles2018}. Here too, we identified environments as the limiting factor; brat~\cite{brat2012}, the annotation tool used to annotate action graphs, was suited for sentence-level, span-based annotation, and not for converting texts into process-level and executable code. Using TextLabs we annotated a new dataset of executable protocols, by recording the action sequences of annotators. The text-based game environment enabled collection of comparatively long and detailed action sequences, while capturing complex phenomena such as long range co-reference, common-sense reasoning and implicit arguments. We used the data to develop graph-prediction models, finding them to be good at entity identification and local relation extraction, yet challenged by long-range relations.

\subsubsection{Evaluating and enhancing world-modelling capabilities of NLU systems}

Chapters \ref{chap:dynababi} and \ref{chap:breakpoint_transformers} explored questions related to world-modelling capabilities in neural language models. World-modelling (also called situation modelling) refers to the ability to construct a model of a situation described in natural language. It requires common-sense knowledge about agents and objects as well as reasoning about the effects of events over time. World-modelling is a classic example of an ``environment bottleneck:'' despite its importance as a cornerstone of human cognition (Chapter \ref{chap:lang_remodel}), there is a notable lack of NLU benchmarks for rigorously evaluating it. World-modelling benchmarks are challenging to create as they typically require a micro-world simulator environment that is both complex enough to construct interesting tasks, and controllable enough to facilitate precise experimentation. The nearest existing benchmark, bAbI~\cite{babi2016}, provided only limited task complexity and controllability. In Chapter \ref{chap:dynababi}, we  developed a new task generator called Dyna-bAbI, and used it to create bAbI 2.0, a challenging new suite of tasks, with a particular focus on compositional generalization, an important evaluation setting absent from the original benchmark. We evaluated a wide array of models including both specialized architectures developed for bAbI, as well as general purpose pre-trained models such as RoBERTa~\cite{roberta2019} and T5~\cite{2020t5}. We found that while pre-trained models far outperformed specialized models, neither class of model performed well on the compositional generalization tasks. These results indicate the limitations of those models as well as the data included in the original benchmark. We explored ways to augment the original data, and found that though diversifying training data was far more useful than simply increasing dataset size, it was still insufficient for driving robust compositional generalization. We also evaluated our new benchmark with a newly developed quality metric called concurrence~\cite{liu2023question}. We found that bAbI 2.0 significantly outperformed the original bAbI dataset and achieved concurrence results on par with purpose-built high concurrence synthetic benchmarks, as well as widely used natural language benchmarks. Our results underscore the importance of highly controllable task generators for creating robust NLU systems through a virtuous cycle of model and data development.\\

Chapter \ref{chap:breakpoint_transformers} leveraged the Dyna-bAbI task generator to improve the world-modelling capabilities of existing NLU models. Drawing inspiration from the concept of breakpoints in programming, we developed a new approach called Breakpoint Modelling that enables models to learn representations of semantic information at intermediate points throughout long-form text. Breakpoints are simply special tokens (similar to the widely used [CLS] tokens) inserted after each sentence, and the resulting encoded breakpoint vector $b$ can then be queried against a natural language proposition vector encoding $p$ to obtain a true/false/unknown prediction for each $\left(b,p\right)$ pair. Our original goal was to explore whether providing dense world-state annotations as supervision would improve compositional generalization performance on bAbI 2.0. Though we observed only marginal gains on compositional generalization tasks, breakpoints proved valuable for interpretability, e.g., understanding model behavior and anticipating potential failure modes. More importantly, we found that the breakpoint modelling training objective can be added to the sequence to sequence loss as used in state-of-the-art Transformer models like T5. Such models can be trained jointly to predict intermediate propositions alongside any existing capabilities like question answering, without harming their original performance. This means that the resulting model, which we called a Breakpoint Transformer, can answer questions and efficiently predict hundreds of propositions in a single forward pass, as opposed to baseline methods which can typically only predict one breakpoint-proposition pair per forward pass. We applied Breakpoint Transformers to the TRIP benchmark~\cite{storks2021tiered}, a multi-step reasoning task evaluating world-modelling capabilities using short stories with reasoning ``bugs'' or implausible sequences of events (e.g., a telephone rang after it had been unplugged). Our model obtained state-of-the-art performance on TRIP, including 20-30\% absolute improvement on 2 out of the 3 sub-tasks. The TRIP experiments additionally demonstrated the versatility of our approach: we converted the full 3-task pipeline to breakpoint modelling format (proposition prediction and question answering) without further architectural changes,  whereas the baseline approach involved task-specific architectural adaptations to a RoBERTa model.

\section{Environments for AI-augmented collective human thinking}
\label{sec:eco_sense}


 \begin{quotation}
\noindent ``If you want to teach people a new way of thinking, don't bother trying to teach them. Instead, give them a tool, the use of which will lead to new ways of thinking.'' -- R. Buckminster Fuller
 \end{quotation}

Chapter \ref{chap:stigmergic_annotation}, the final chapter of the thesis, was, in many respects, written in a very different world than the one from which I set out on my thesis journey in 2018. Outside the window of my lab the world was quite literally burning, rocked by climate catastrophes, a global pandemic, and rapidly deteriorating social cohesion world-wide and in my home country of Israel. Rampant misinformation and information weaponization highlighted the perils of outsourcing crucial societal communication infrastructure to centralized social media platforms with little scientific oversight, and monopolistic control of data and algorithms. If indeed environments play such an integral role in human (and machine) cognition, perhaps many of those “epistemic ills” plaguing societies (e.g., fragmentation, misinformation) today can be traced back to the online social media environments from which humans increasingly acquire their information about the world? In Chapter \ref{chap:stigmergic_annotation} we ask, what would healthier epistemic environments look like, that support the learning and sensemaking journeys of individuals and collectives? How to create environments in which AI would serve to augment cognition rather than manipulate it for ad-driven engagement? Our proposal hinges on the concept of \emph{stigmergy} which originated from studies of collective intelligence in ant colonies~\cite{Marsh2008}. Stigmergy is the phenomena of large scale coordination mechanism mediated by local environment modifications (e.g., ant pheromone trails). Theories of stigmergic cognition foreground the role of the environment, which functions as a distributed memory system for a collective organism. Annotations also play a pivotal role in stigmergy, as they function as markers of meaning (e.g., annotations such as likes on social media are ``digital pheromone trails''). We applied principles of stigmergy to propose a new kind of social network designed to address the limitations of centralized platforms: an open, decentralized social annotation network where participants control their annotations, and can share them with AI-driven content discovery services. Decentralization provides resilience against platform capture, and open data encourages a plurality of content discovery services, as opposed to platforms' monolithic feed algorithm.

\chapter{Language (Re)modelling: Towards Embodied Language Understanding
}
\label{chap:lang_remodel}
{\Large Ronen Tamari, Chen Shani, Tom Hope, Miriam Petruck, Omri Abend, Dafna Shahaf}
\\\\
\noindent {\large Published in the Annual Meeting of the Association for Computational Linguistics (ACL), 2020}
\newpage
\includepdf[pages=-]{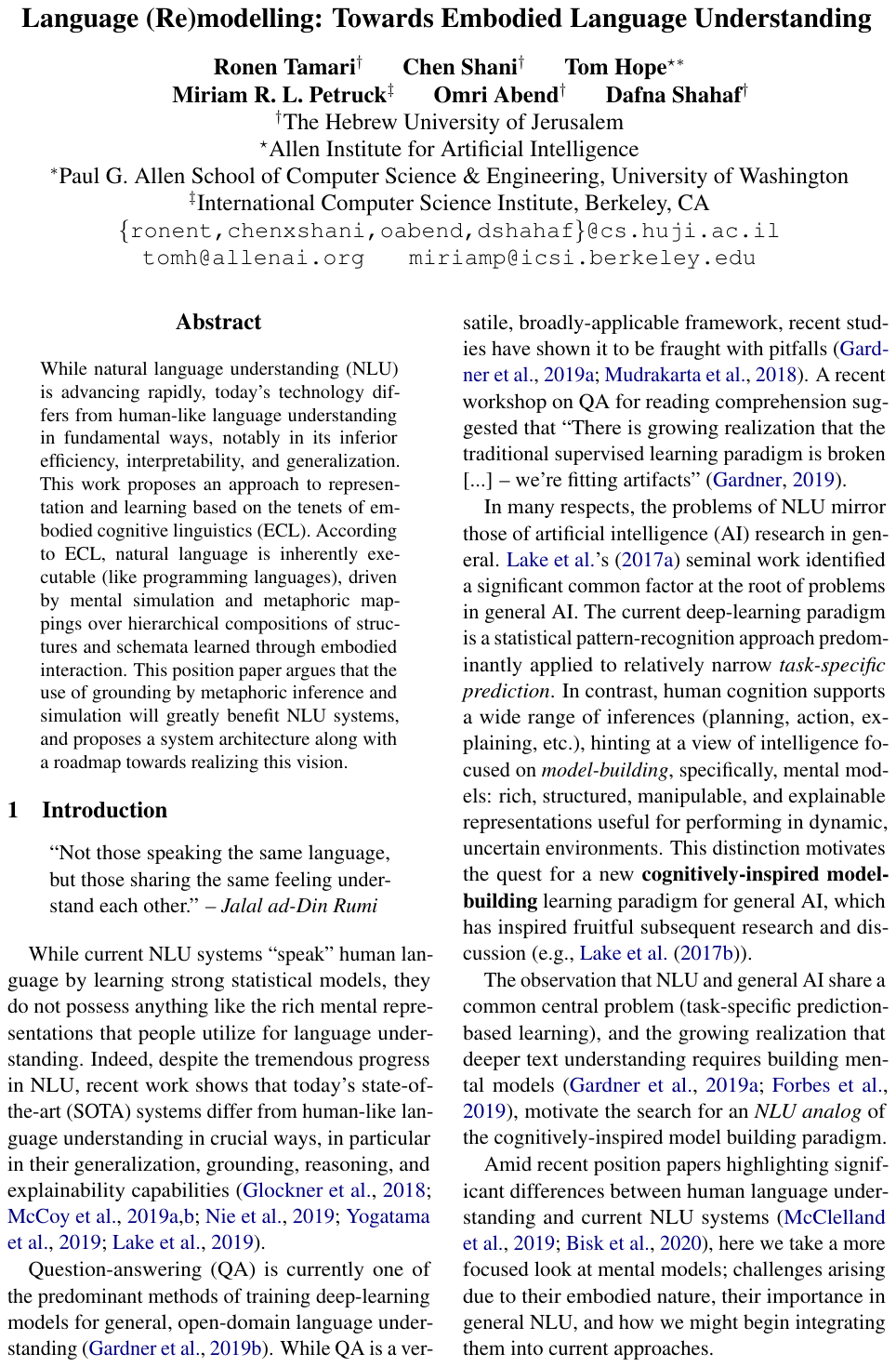}

\chapter{Playing by the Book: An Interactive Game Approach for Action Graph Extraction from Text
}
\label{chap:ptb}
{\Large Ronen Tamari, Hiroyuki Shindo, Dafna Shahaf, Yuji Matsumoto}
\\\\
\noindent {\large Published in the Proceedings of the Workshop on Extracting Structured Knowledge from Scientific Publications (ESSP), 2019}
\newpage
\includepdf[pages=-]{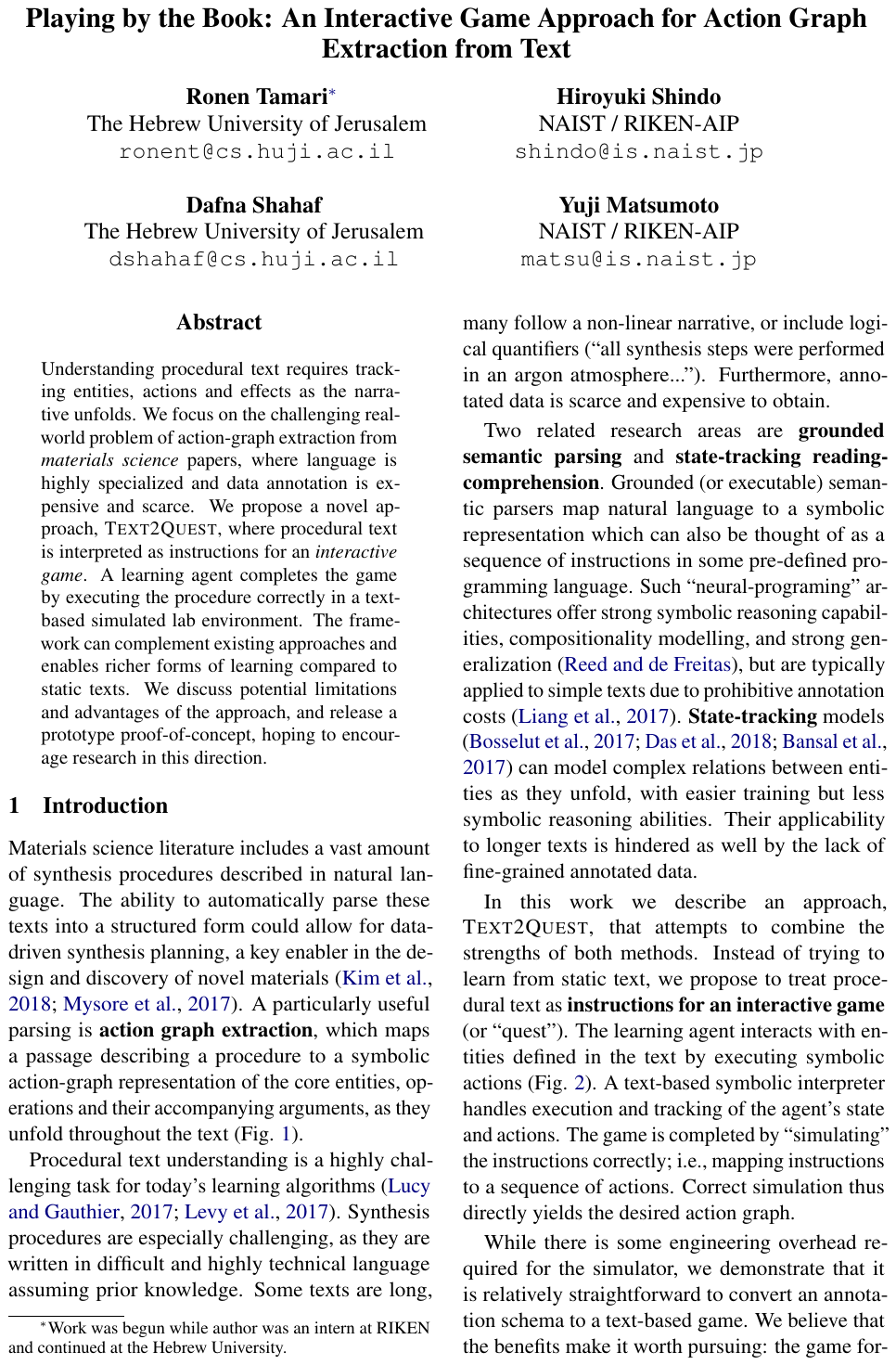}

\chapter{Process-Level Representation of Scientific Protocols with Interactive Annotation}
\label{chap:scientific_protocols}
{\Large Ronen Tamari, Fan Bai, Alan Ritter, Gabriel Stanovsky}
\\\\
\noindent {\large Published in the European Chapter of the Association for Computational Linguistics (EACL), 2021}
\includepdf[pages=-]{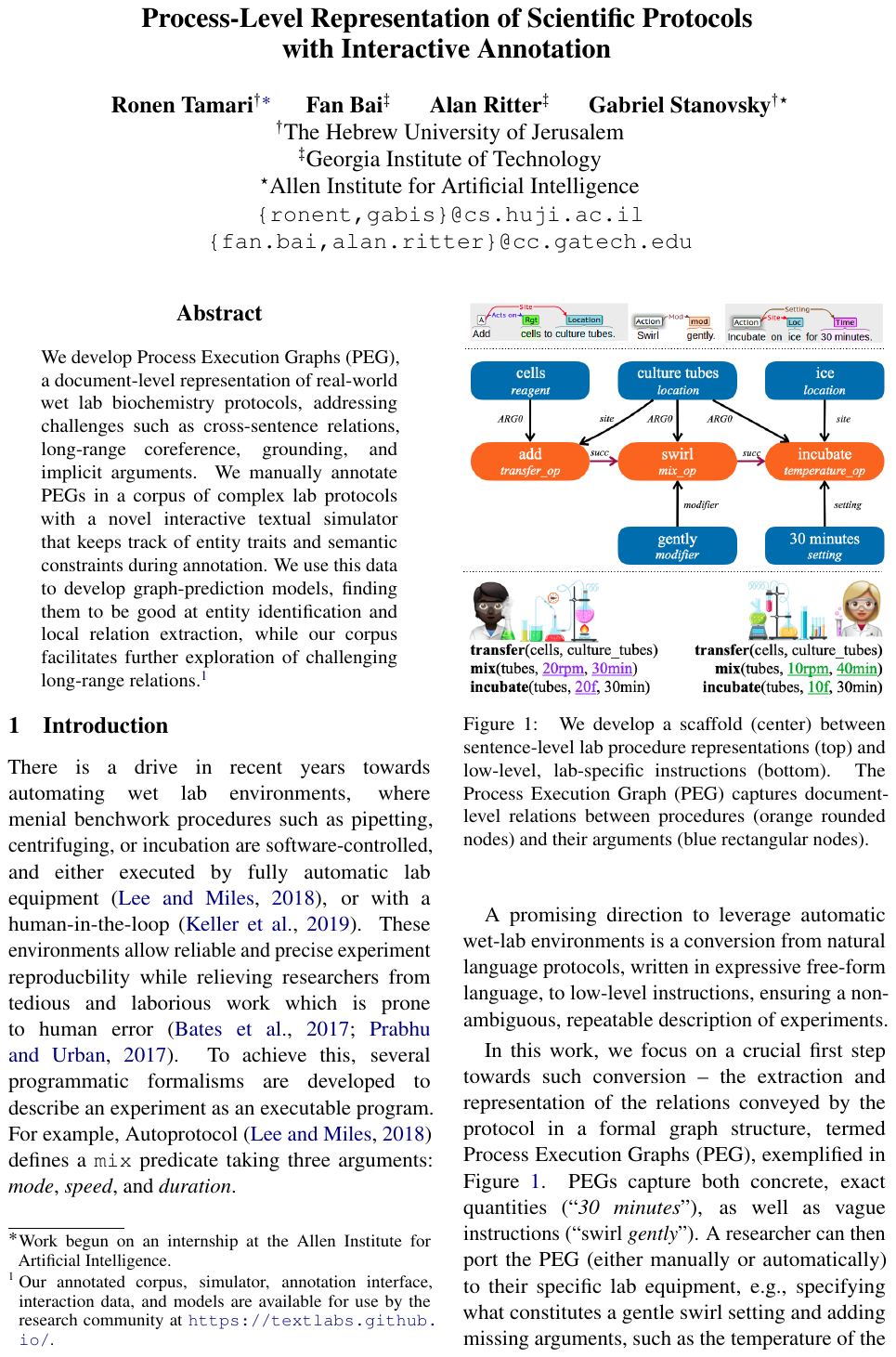}

\chapter{Dyna-bAbI: unlocking bAbI's potential with dynamic synthetic benchmarking}
\label{chap:dynababi}
{\Large Ronen Tamari, Kyle Richardson, Aviad Sar-Shalom, Noam Kahlon, Nelson Liu, Reut Tsarfaty, Dafna Shahaf}
\\\\
\noindent {\large Published in the Joint Conference on Lexical and Computational Semantics (*SEM), 2022}
\includepdf[pages=-]{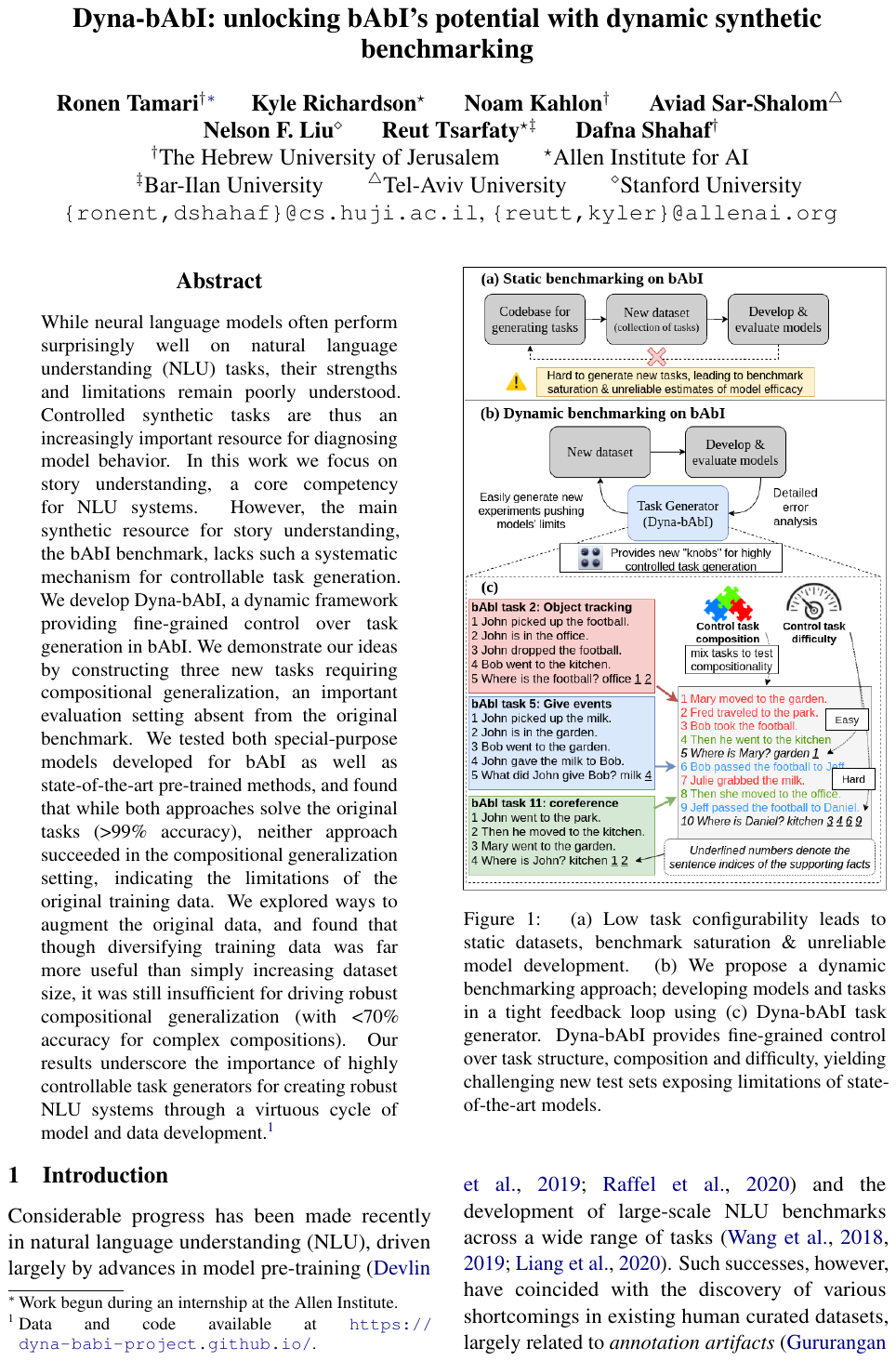}

\chapter{Breakpoint Transformers for Modeling and Tracking Intermediate Beliefs}
\label{chap:breakpoint_transformers}
{\Large Kyle Richardson$^{*}$, Ronen Tamari$^{*}$, Oren Sultan, Dafna Shahaf, Reut Tsarfaty, Ashish Sabharwal}
\\\\
\noindent {\large Published in the Conference on Empirical Methods in Natural Language Processing (EMNLP), 2022}
\includepdf[pages=-]{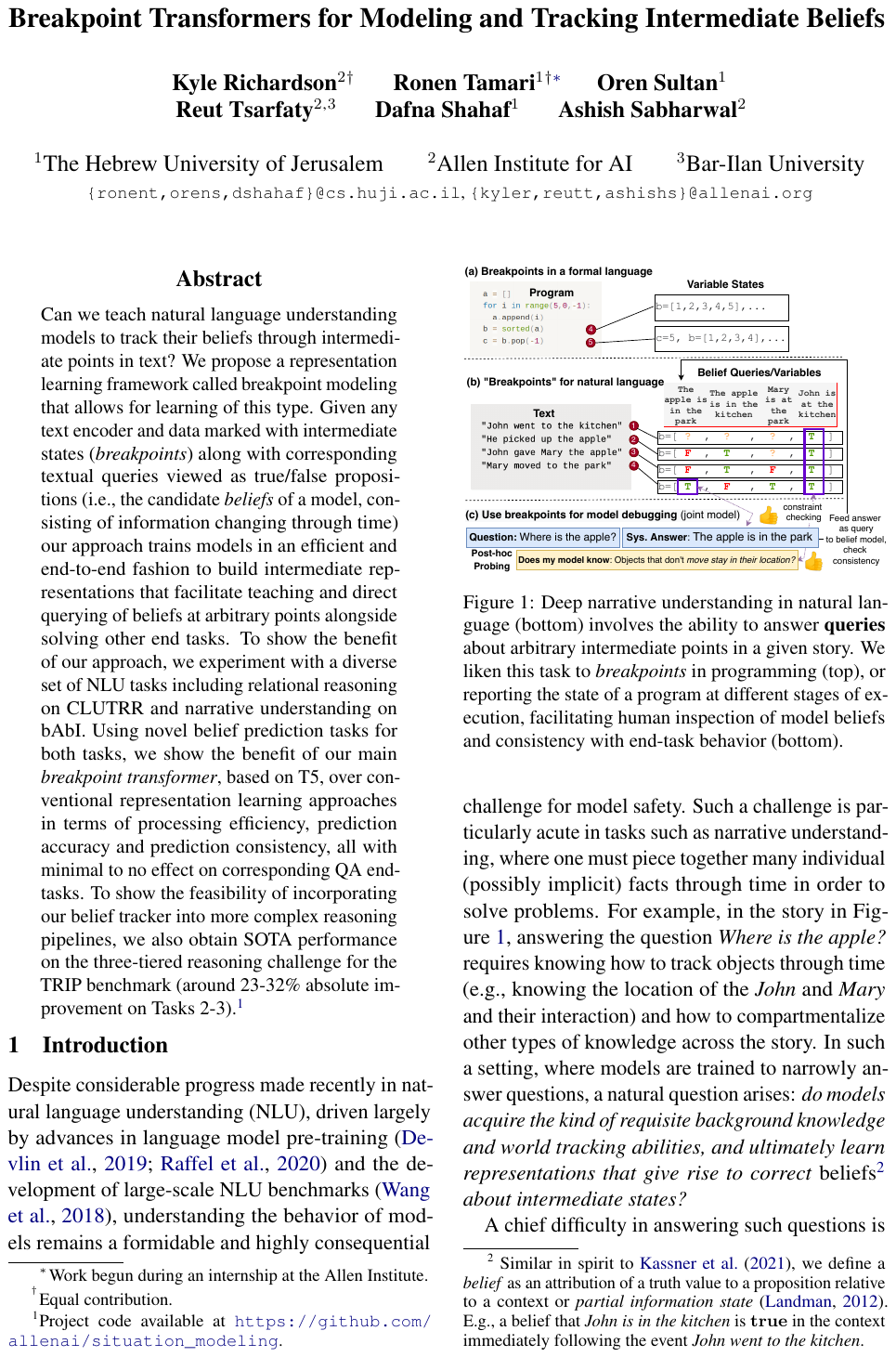}

\chapter{From Users to (Sense)Makers: On the Pivotal Role of Stigmergic Social Annotation in the Quest for Collective Sensemaking}
\label{chap:stigmergic_annotation}
{\Large Ronen Tamari, Daniel Friedman, William Fischer, Lauren Hebert, Dafna Shahaf}
\\\\
\noindent {\large Published in the ACM Conference on Hypertext and Social Media (HT), 2022}
\includepdf[pages=-]{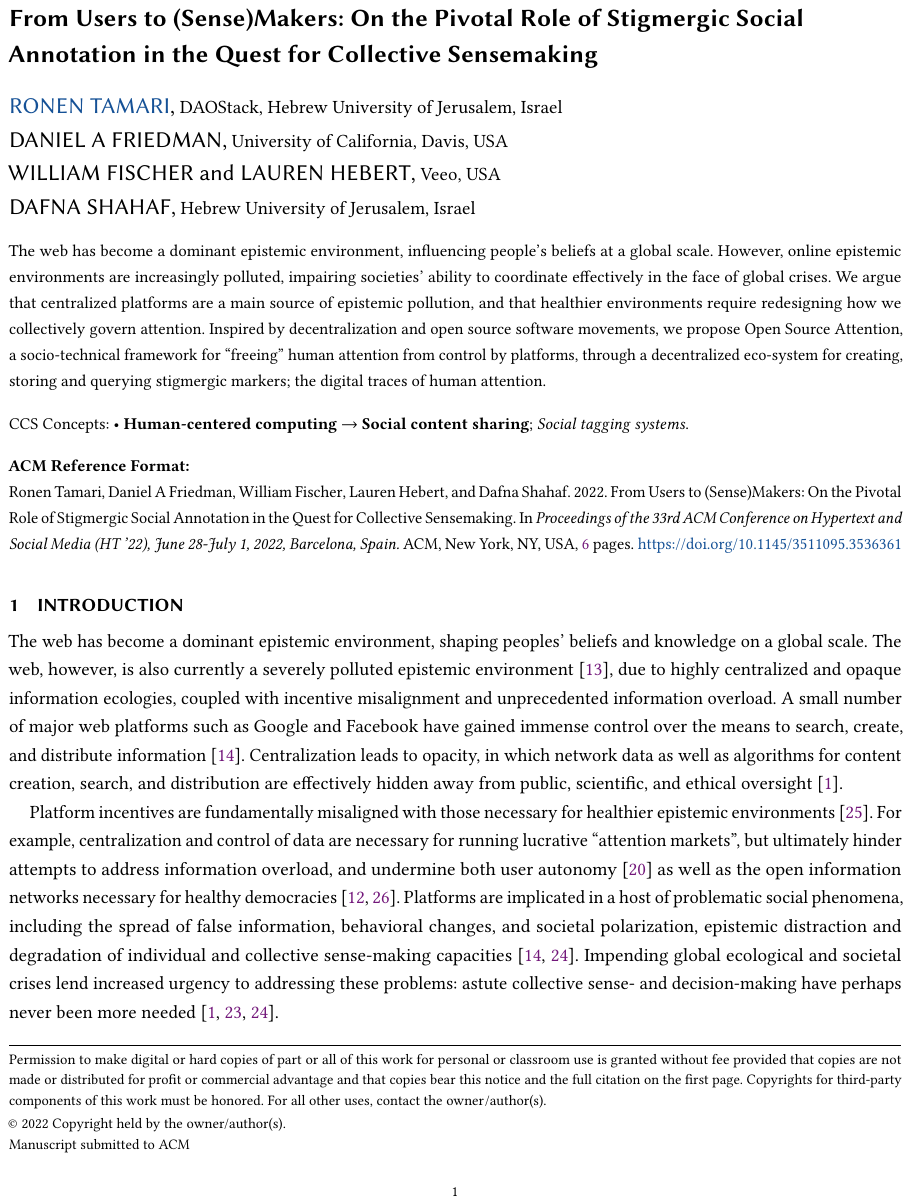}

\chapter{Discussion and Conclusions}
\label{chap:conclusion}

 \begin{quotation}
\noindent ``The computer is only an arc of a larger circuit which always includes a man and an environment'' -- Gregory Bateson
 \end{quotation}

In this work, we explored environments in NLU research, both in theory as well as their role as technical infrastructure supporting model training and evaluation, and data collection and annotation. We developed an \emph{ecological NLU} approach foregrounding the role of environments. Our thesis was that (1) environments are a crucial enabler of progress both on conceptual and empirical fronts in NLU. (2) Environments receive relatively little attention compared with modelling (architecture, optimization, learning algorithms). As a result, many research questions are hindered by \emph{environment bottlenecks} in which relevant frameworks or experiments cannot be implemented due to the lack of suitable environments. 

The research in this thesis spans five years, a very long time in contemporary NLU which has been turbo-charged by large-scale industry-based resource investment. As a result, at the conclusion of the thesis we have the chance to look back and see how our research has aged, and discuss future extensions of our work.

In Chapter \ref{chap:lang_remodel} we identified an environment bottleneck at the conceptual level; the standard language modelling formulation omits environments, accounting only for distributional meaning that lies in the statistical relations between words. We extended this model to a novel embodied language model, informed by theories of embodied cognitive linguistics, that additionally accounts for the mental states of interlocutors. In this model, language acts upon interlocutors' world models which are learned through interaction with an environment. We proposed situating language in an agent-based context including perceptions, actions and an external world that can be interacted with. How has this proposal fared? On one hand, in the years since we published the work, rapid progress on pre-trained large language models has demonstrated just how much linguistic meaning can be gleaned from statistical patterns alone, given enough data and parameters. Few researchers, ourselves included, expected the emergence of reasoning capabilities and world knowledge to the degree displayed by the latest models such as GPT-4. That said, our framework predicts models will fare poorly on language requiring the tracking of dynamic world states (world-modelling), and compositional reasoning. Rigorous evaluation indicates that this remains the case for newer models~\cite{dziri2023faith} as well as the previous generation of LLMs tested in Chapter \ref{chap:dynababi}. Models like Chat-GPT still fail to solve even the \emph{original} bAbI benchmark reliably~\cite{xiang2023language}. Finally, agent-based language models are a promising and rapidly evolving field of research~\cite{qiu-etal-2022-towards,wang2023interactive,sumers2023cognitive}, highlighting the relevance of our proposal. Future work could systematically benchmark such models on bAbI 2.0 (Chapter \ref{chap:dynababi}) or extensions of it. Much work remains to better understand the nature of the implicit or vector world models such as we explored in Chapter \ref{chap:breakpoint_transformers}. Promising approaches include the incorporation of neuro-symbolic learning algorithms to promote logic consistency~\cite{li2019logic}, as well as more systematic probing into the nature of learned world models~\cite{nanda2023emergent}.

In Chapters \ref{chap:ptb} \& \ref{chap:scientific_protocols}, we identified an environment bottleneck in the annotation and representation methods used in NLU systems for scientific experimental protocols. Prior frameworks enabled span-level annotation but did not support applications requiring deeper process-level comprehension, such as text-to-code for parsing instructions to machine executable programs in cloud laboratories. Support for such applications necessitates new annotation methods, domain specific languages, and execution environments. We showed how text-based games could be leveraged as inexpensive lab simulators supporting annotation, richer interactive training, and synthetic data generation for process-level applications. To our knowledge, our works were among the first to propose a framework for parsing natural language experimental protocols to executable code. Recent years have seen increased interest in this setting across biology, chemistry and materials science~\cite{Wang_Cruse_2022,Park_Mill_Arrechea_2023,Zeng_2023}. The ChemCrow system~\cite{bran2023chemcrow} demonstrated the power of agent-based LLMs coupled with domain-specific interactive chemistry environments, or tools: the tool-augmented GPT-4 far out-performed a baseline GPT-4 model across a variety of tasks such as chemical synthesis and material design. Using models to code interactive environments was proposed by us as a promising future direction in Appendix \ref{chap:eco_sem} and was recently demonstrated as a viable, though still highly challenging possibility~\cite{wang2023bytesized32}.  Another promising future direction involves LLM agents learning through interaction with environments, as opposed to frozen LLMs being prompted to call APIs in an iterative setting~\cite{schick2023toolformer}. Training multi-modal interactive LLMs~\cite{lin2023learning} is an intriguing approach to the important open question of whether embodied experience may improve language models' (linguistic) world-modelling capabilities, as discussed in Chapter \ref{chap:lang_remodel}.  Finally, as discussed briefly in 
Appendix \ref{chap:eco_sem}, environment research would benefit enormously from more powerful computational libraries to support coding environments, similar to the pivotal role played by PyTorch and Tensorflow for modelling research.

\subsubsection{The understated role of environments in the success of LLMs}
The impressive performance of LLMs, while challenging prior assumptions on the limitations of distributional NLU approaches, at the same time also validate the central premise of ecolgical NLU that the importance of modelling tends to be overrated, and that environments matter far more than previously expected. This can be seen from the relative simplicity of current LLM architecture and training methods, which for the most part are Transformer models~\cite{Vaswani_2017} pre-trained on next-word prediction tasks, with multi-head attention as employed in the GPT line of models~\cite{Radford2019LanguageMA}. Subsequent modelling breakthroughs have been relatively much less important than simply scaling parameters and dataset size and diversity. Lifting the hood on dataset size and diversity, we find: environments. Platforms like Reddit and Wikipedia form the backbone of LLM training datasets, but also happen to be diverse, carefully constructed online environments to which humans have collectively contributed a vast array of linguistic knowledge. We'll try to make the point more concrete with a thought experiment: imagine a parallel, pre-LLM world (circa 2018), in which Stack Overflow and GitHub do not exist. In this world, a group of AI researchers wants to develop an AI coding assistant that can understand code and help write it. How would they go about solving the problem? The group would perhaps consider recruiting experts to craft a dataset of questions and answers, and would likely focus a lot of attention on modelling: elaborate neuro-symbolic architectures, relational reasoning modules, graph-based representations, etc. They would in all likelihood \emph{not} undertake what we know with hindsight to be the most critical move, which is, \emph{to build platforms like Stack Overflow and Github}.  Those platforms would collect the quality human interactions at the scale necessary for training a relatively simple pre-trained coding assistant LLM like Codex~\cite{chen2021evaluating}. The challenges involved in building that platform would be human and engineering challenges, not modelling challenges; incentives, UX, scale, moderation, and so on. For this reason, that route would be overlooked by AI researchers, who are experts in building models, not environments. While this story may be somewhat contrived, the question still stands: how many other AI or NLU problems that we as a community are trying to solve with fancy new modelling approaches, are actually more about building the right human interaction environments?

\subsubsection{Environments and models matter for society}

Chapter \ref{chap:stigmergic_annotation} raises the future directions I personally am most excited about, as it draws on that ``environment vs model'' question and applies it to real-world societal challenges of misinformation, polarization and information overload. Big tech companies would like us to think that their corporate LLM conversational agents are a one-stop shop for our epistemic needs, but many serious warnings have been raised about these model-centric solutions and their harms~\cite{Shah2022Situating,Saunders_2023}. We asked instead, what is the ``Github of collective sensemaking'' that will support the meaningful human interactions that will power future sensemaking assistants? We found another environment bottleneck: no such environments currently exist. The nearest approximation can be found in social media platforms such as Twitter, which, while demonstrating the occasional brilliance of AI-augmented collective human intelligence, also reveal the perils of centralization and opaque data and algorithms. We outlined a more human-centered, decentralized approach to collective sensemaking: we proposed empowering people with semantic web annotation tools to mark, review and link between content, and using AI and social trust graphs to leverage relevant “digital trails” left by others.

In summary, this thesis began with a search for better NLU models and is ending as the start of a new search for better environments. With data-driven algorithmic models rapidly gaining influence on human societies across all spheres of life, seeing the often invisible environments at the nexus of humans, models and data has perhaps never been more urgent. It is my hope that this thesis will encourage other AI researchers to step back and take the ecological perspective, or, to paraphrase Mace: to ``ask not what's inside your model, but what your model's inside of.''

\bibliographystyle{ieeetr}
\bibliography{egbib}

\clearpage
\newpage
\appendix
\chapter{Ecological Semantics: Programming Environments for Situated Language Understanding}
\label{chap:eco_sem}
{\Large Ronen Tamari, Gabriel Stanovsky, Dafna Shahaf, Reut Tsarfaty}
\\\\
\noindent {\large Preprint published electronically, 2020. (DOI: 10.48550/arXiv.2003.04567)}
\newpage
\includepdf[pages=-]{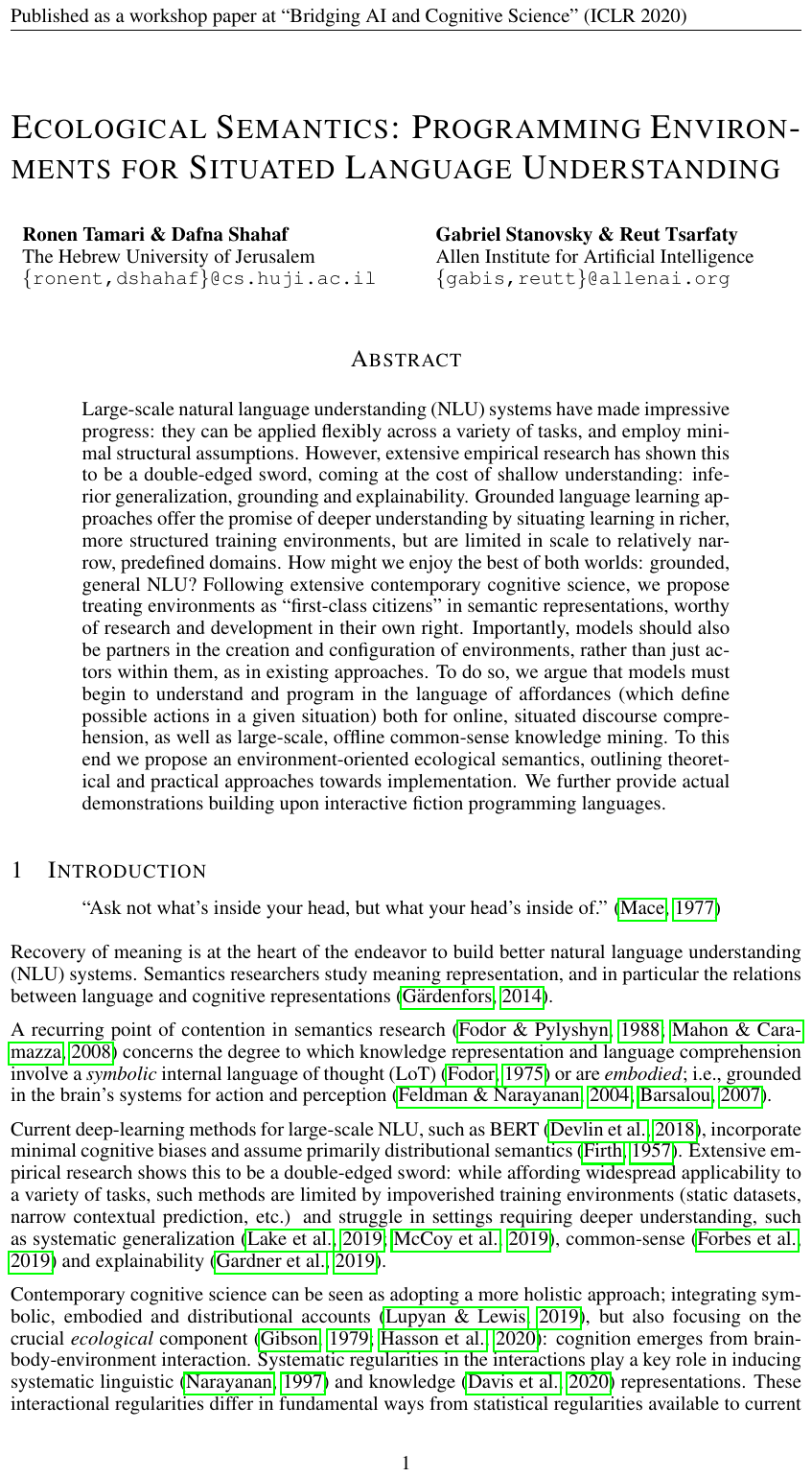}

\mbox{}

\end{document}